\documentclass{article}
\usepackage{epsfig,fullpage,psfrag,url}
\usepackage{graphics}
\usepackage{graphicx}
\usepackage{amsmath, amsthm, amssymb}

\usepackage{amsfonts}
\usepackage{algorithmic}
\usepackage{algorithm}

\usepackage{color}

\newtheorem{thm}{Theorem} 
\newtheorem{lem}{Lemma}

\newcommand{\s}{\mathbf S}

\newcommand{\half}{\mbox{$\frac12$}}
\newcommand{\rnk}{\mathrm{rank}}

\newcommand{\sbt}{\mathrm{subject\;\;to}}
\DeclareMathOperator*{\argmin}{arg\,min}
\DeclareMathOperator*{\mini}{minimize}

\title{Regularization methods for learning incomplete matrices}
\author{
Rahul Mazumder 
 \thanks{
 Statistics Department,Stanford University
 \texttt{rahul.mazumder@gmail.com}}\\
\and
Trevor Hastie \thanks{Statistics Department and Department of Health, Research and Policy,
Stanford University, 
\texttt{hastie@stanford.edu}}
\and
 Robert Tibshirani
\thanks{
Department of  Health, Research and Policy and Statistics Department, Stanford University
 \texttt{tibs@stanford.edu}
 }
}

\begin{document}

\maketitle

\begin{abstract}
  We use convex relaxation techniques to provide a sequence of
  solutions to the matrix completion problem. Using the nuclear norm
  as a regularizer, we provide simple and very efficient algorithms
  for minimizing the reconstruction error subject to a bound on the
  nuclear norm. Our algorithm iteratively replaces the missing
  elements with those obtained from a thresholded SVD. With warm
  starts this allows us to efficiently compute an entire
  regularization path of solutions.
\end{abstract}

\section{Introduction}
In many applications measured data can be represented in a matrix
$X_{m\times n},$ for which only a relatively small number of entries
are observed.  The problem is to ``complete'' the matrix based on the
observed entries, and has been dubbed the matrix completion problem
~\cite{cai-2008,candes:recht,recht-2007,candes-2009,monti-09}.  The
``Netflix'' competition is a primary example, where the data is the basis
for a  recommender
system.  The rows correspond to  viewers and the columns to movies, with
the entry $X_{ij}$ being the rating $\in\{1,\ldots,5\}$ by viewer $i$ for movie
$j$. There are 480K viewers and 18K movies, and hence 8.6 billion
($8.6 \times 10^9$) potential
entries. However, on average each viewer rates about 200 movies, so
only 1.2\% or $10^8$ entries are observed.
The task is to predict the ratings viewers would give for the movies
they have not yet rated.

These problems can be phrased as learning an unknown parameter (a matrix $Z_{m
  \times n}$) with very high dimensionality, based on very few
observations. In order for such inference to be meaningful, we assume
that the parameter $Z$ lies in a much low
dimensional manifold.  In this paper, as is relevant in many real life
applications, we assume that $Z$ can be
well represented by a matrix of low rank, i.e. $Z\approx
V_{mk}G_{kn}$, where $k\ll\min(n,m)$.  In this recommender system
example, low rank structure suggests that movies can be grouped into a
small number of ``genres'', with  $G_{\ell j}$ the relative score for
movie $j$ in genre $\ell$. Viewer $i$ on the other hand has an
affinity $V_{i\ell}$ for genre $\ell$, and hence the modeled score for
viewer $i$ on movie $j$ is the sum $\sum_{\ell=1}^kV_{i\ell}G_{\ell
  j}$ of genre affinities times genre scores. 
Very recently
~\cite{candes:recht,candes-2009,monti-09} showed theoretically that
under certain assumptions on the entries of the matrix, locations and
proportion of unobserved entries, the true underlying matrix can be
recovered within very high accuracy. Typically we view the observed
entries in $X$ as the corresponding entries from $Z$ contaminated with noise. 

For a matrix $X_{m\times n}$ let $\Omega\subset \{1,\ldots,m\}\times
\{1,\ldots,n\}$ denote the indices of observed entries. We consider the
following optimization problem: 
\begin{eqnarray}
\mini && \rnk (Z) \nonumber \\
\sbt && \sum_{(i,j)\in \Omega} (Z_{ij}-X_{ij})^2 \leq \delta,
\label{crit:one}
\end{eqnarray}
where $\delta\geq 0$ is a regularization parameter controlling the
tolerance in training error.  The rank constraint in (\ref{crit:one})
makes the problem  for general $\Omega$
combinatorially hard \cite{Nat-03}.  For a fully-observed $X$, on the
other hand, the solution is given by the singular
value decomposition (SVD) of $X$.  The following seemingly small
modification to (\ref{crit:one})
\begin{eqnarray}
\mini &&  \|Z\|_* \nonumber \\
\sbt && \sum_{(i,j)\in \Omega} (Z_{ij}-X_{ij})^2 \leq \delta
\label{crit:relax}
\end{eqnarray}
makes the problem convex \cite{fazel-thes}.  Here $\|Z\|_*$ is the
nuclear norm, or the sum of the singular values of $Z$. Under many
situations the nuclear norm is an effective convex relaxation to the rank
constraint as explored in
~\cite{fazel-thes,candes:recht,candes-2009,recht-2007}.  Optimization
of (\ref{crit:relax}) is a semi-definite programming problem
\cite{BV2004,fazel-thes} and can be solved efficiently for small
problems, using modern convex optimization software like SeDuMi and
SDPT3. However, since these algorithms are based on second order methods
\cite{int-point}, the problems become prohibitively expensive if the
dimensions of the matrix exceeds a hundred \cite{cai-2008}. In this paper we
propose an algorithm that scales to large problems with $m, n\approx
10^4$--$10^5$ or even larger. We obtain a rank-11 solution to
(\ref{crit:relax}) for a problem of size $(5\times 10^5)\times (5\times
10^5)$ and $|\Omega|=10^4$ observed entries in under 11 minutes
in MATLAB. For the  same sized matrix with
$|\Omega|=10^5$ we obtain a rank-$52$ solution in under 80 minutes.

\cite{candes-2009,cai-2008,candes:recht} consider the criterion
\begin{eqnarray}
\mini &&  \|Z\|_* \nonumber \\
\sbt && Z_{ij}=X_{ij}, \; \forall (i,j)\in \Omega 
\label{crit:candes}
\end{eqnarray}
When $\delta=0$, criterion (\ref{crit:one}) is equivalent to
(\ref{crit:candes}), in that it requires the training error to be
zero.
  \cite{candes-2009,candes:recht}
further develop theoretical properties establishing the equivalence of
the rank minimization and the nuclear norm minimization problems
(\ref{crit:one},\ref{crit:candes}). Cai et. al. ~\cite{cai-2008} in
their paper propose a first-order singular-value-thresholding
algorithm scalable to large matrices for the problem
(\ref{crit:relax}) with $\delta=0.$ They comment on the problem
(\ref{crit:relax}), with $\delta>0$, and suggest that it becomes
prohibitive for large scale problems. Hence they consider the
$\delta>0$ case to be unsuitable for matrix completion.

\begin{figure}[htp]
\centering 
Type a \hspace{3mm}      $50\%$ missing entries with SNR=1, true rank =10\\
\begin{psfrags}
\psfrag{SNR1}[][b]{\small{Test error}}
\psfrag{TrainMiss0.5}[][b]{\small{Training error}}
        \includegraphics[width=3.2in,height=2in]{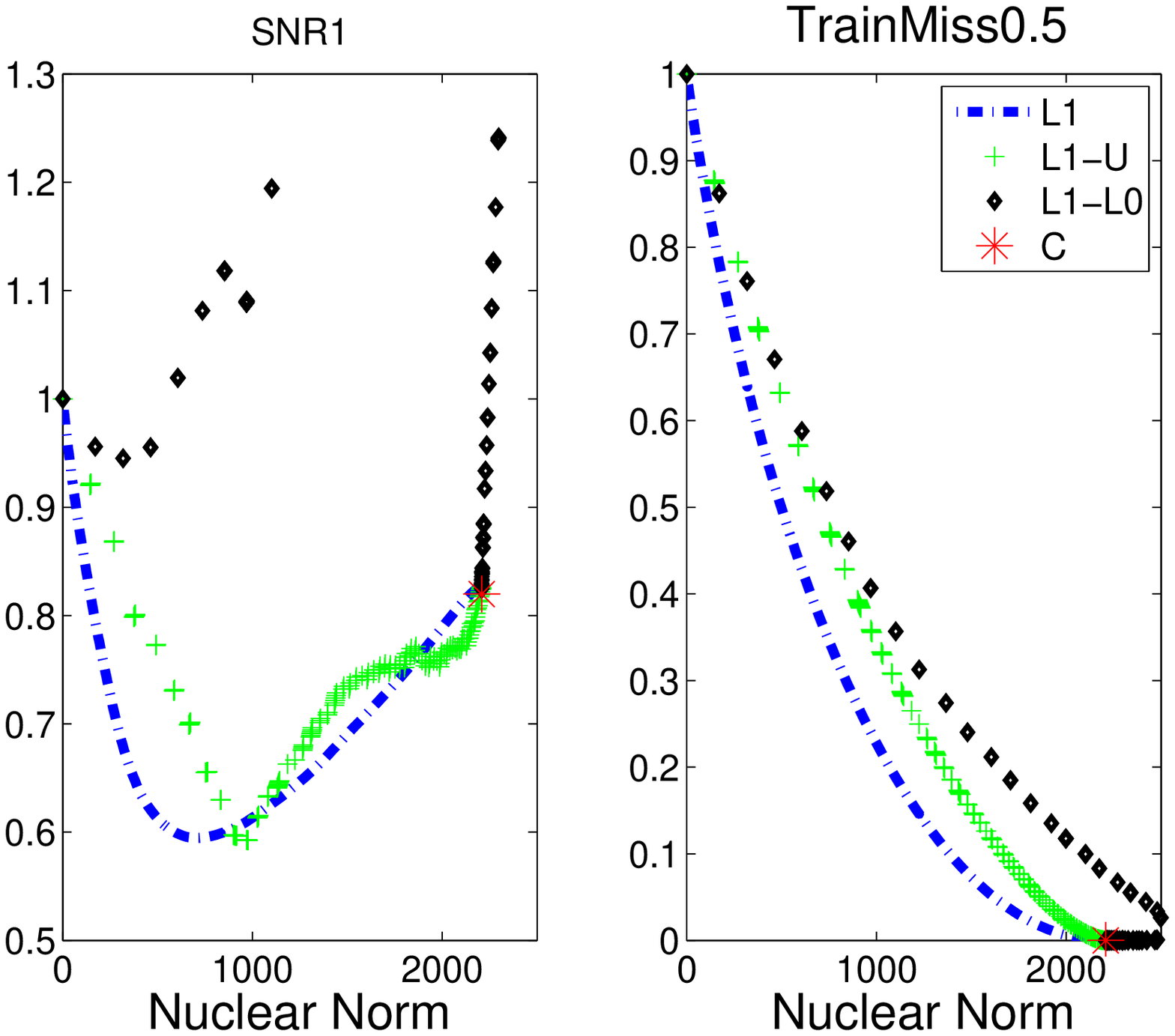} 
	\includegraphics[width=3.2in,height=2in]{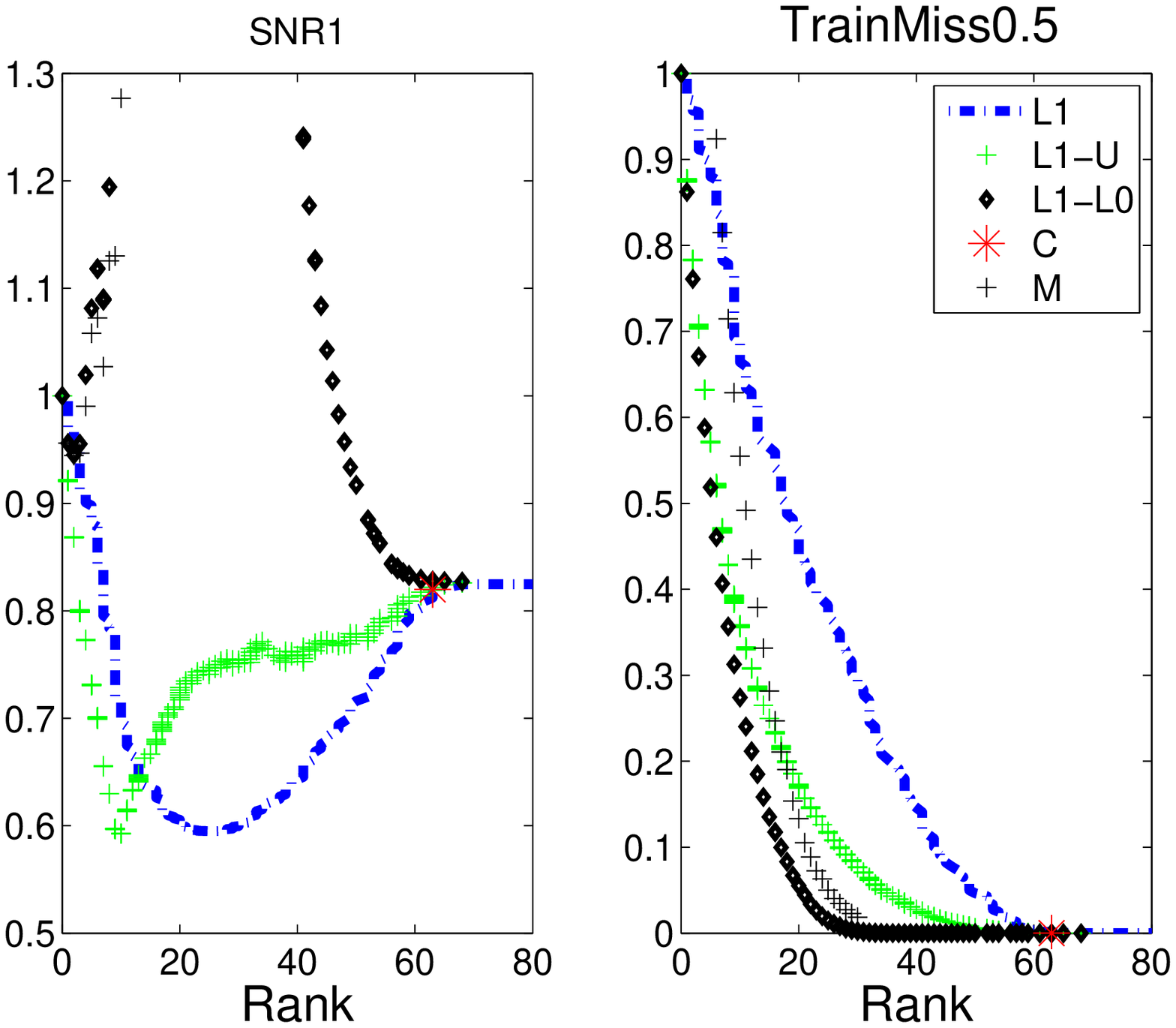} 
\end{psfrags}\\ \vspace{.5cm}
Type b \hspace{3mm}   $50\%$ missing entries with SNR=1, true rank =6\\
\begin{psfrags}
\psfrag{SNR1}[][b]{\small{Test error}}
\psfrag{TrainMiss0.5}[][b]{\small{Training error}}
\includegraphics[width=3.2in,height=2in]{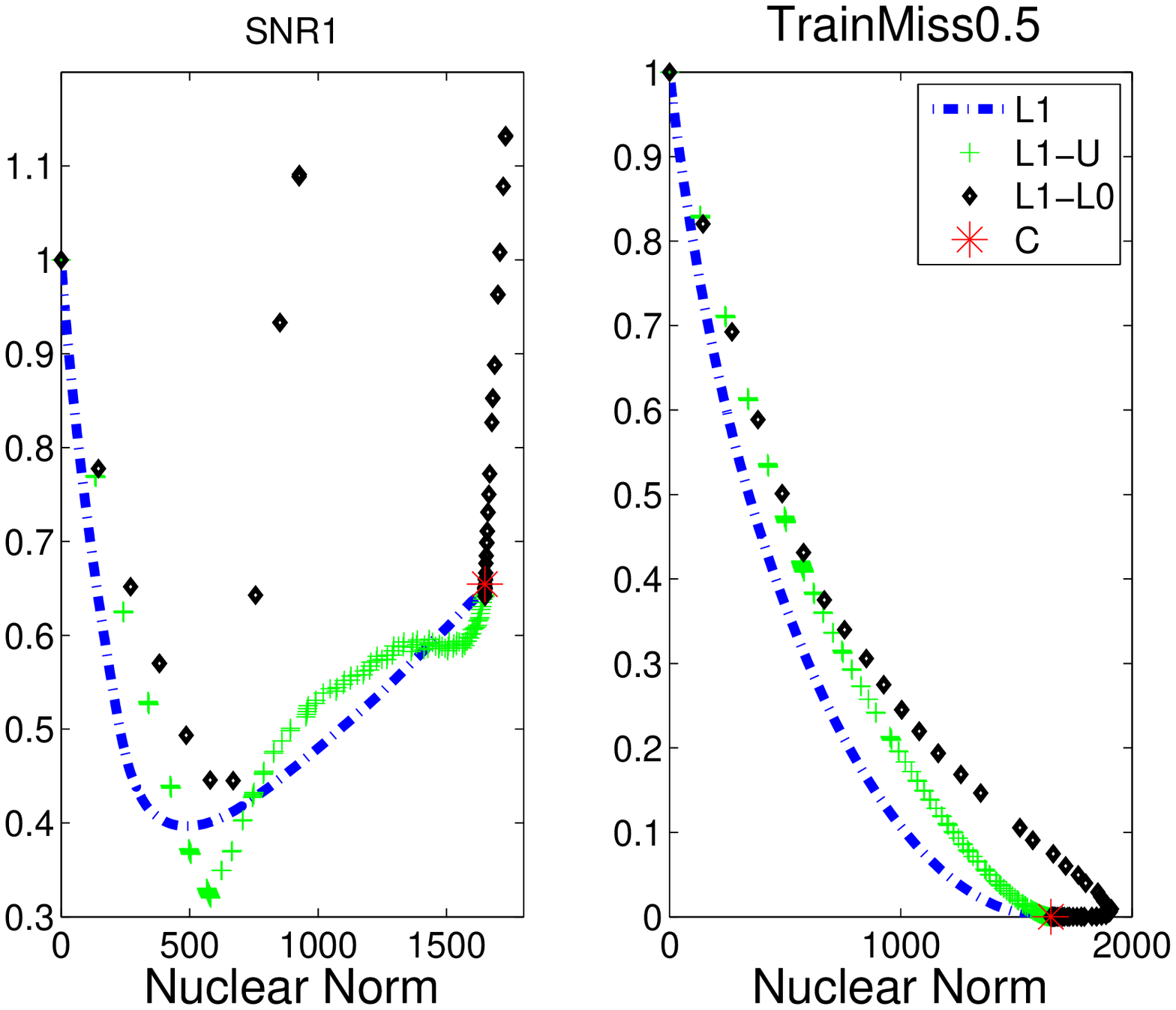} 
	\includegraphics[width=3.2in,height=2in]{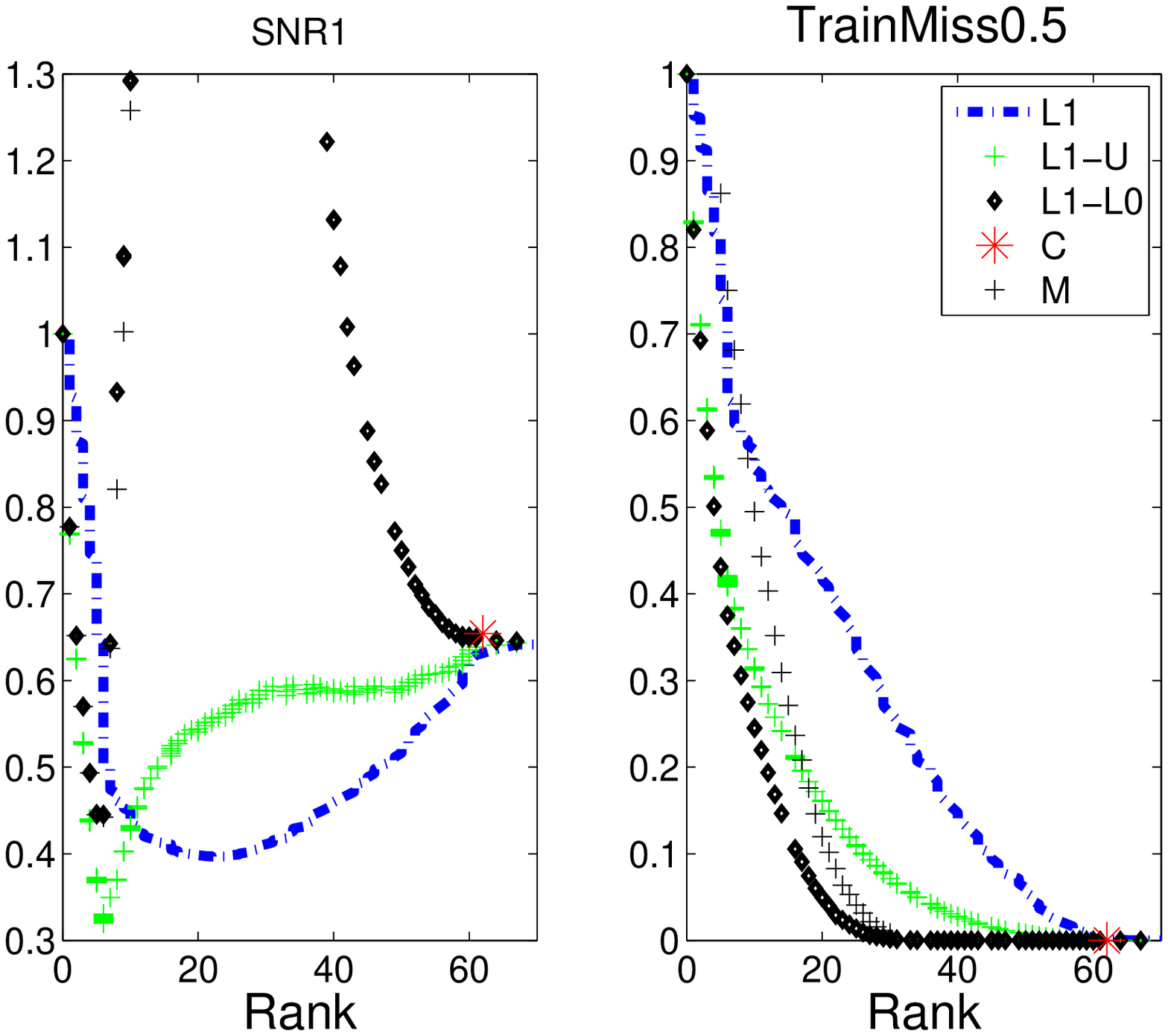} 
\end{psfrags}\\\vspace{.5cm}
Type c \hspace{3mm}   $80\%$ missing entries with SNR=10, true rank =5\\
\begin{psfrags}
\psfrag{SNR10}[][b]{\small{Test error}}
\psfrag{TrainMiss0.8}[][b]{\small{Training error}}
\includegraphics[width=3.2in,height=2in]{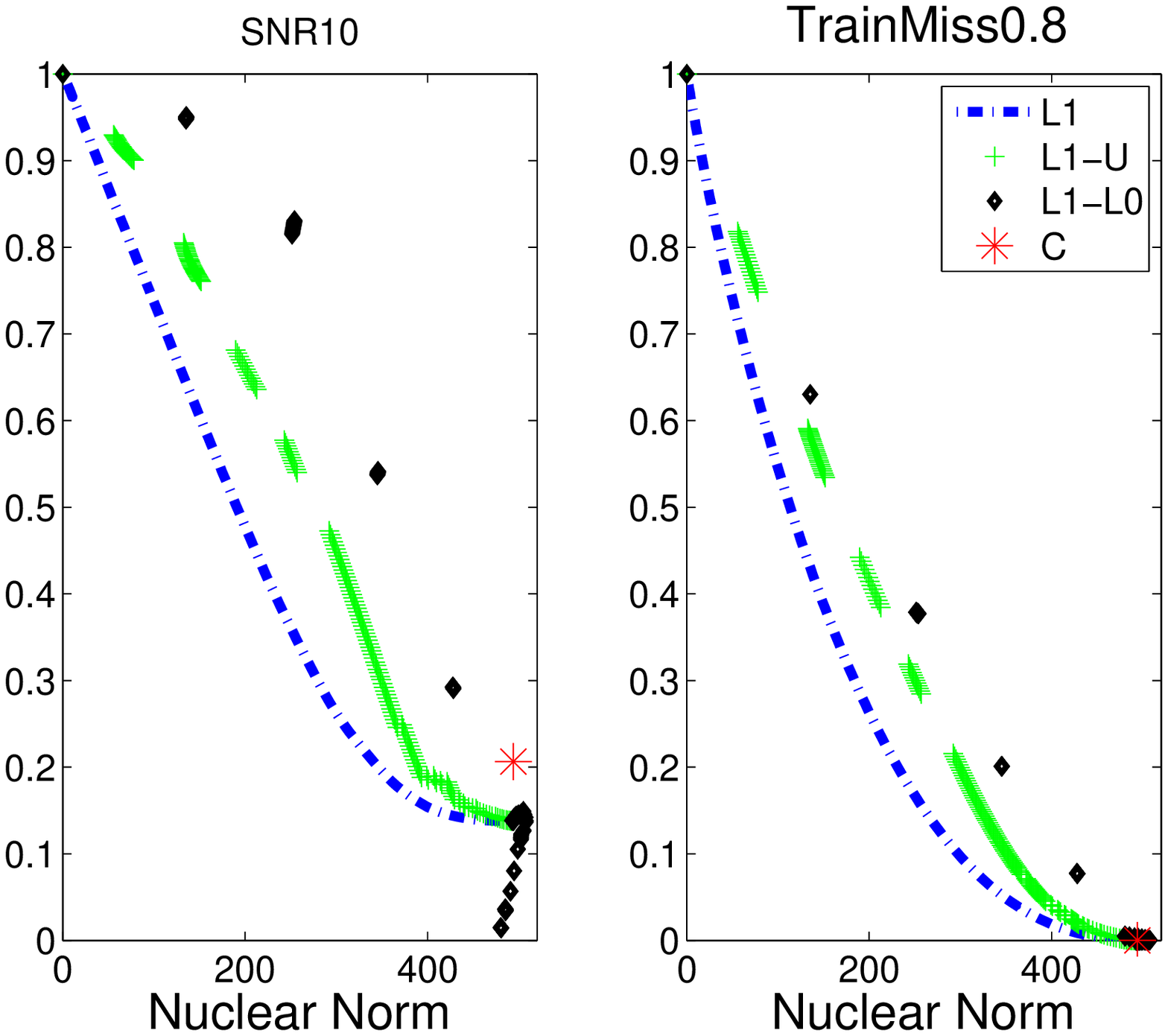} 
	\includegraphics[width=3.2in,height=2in]{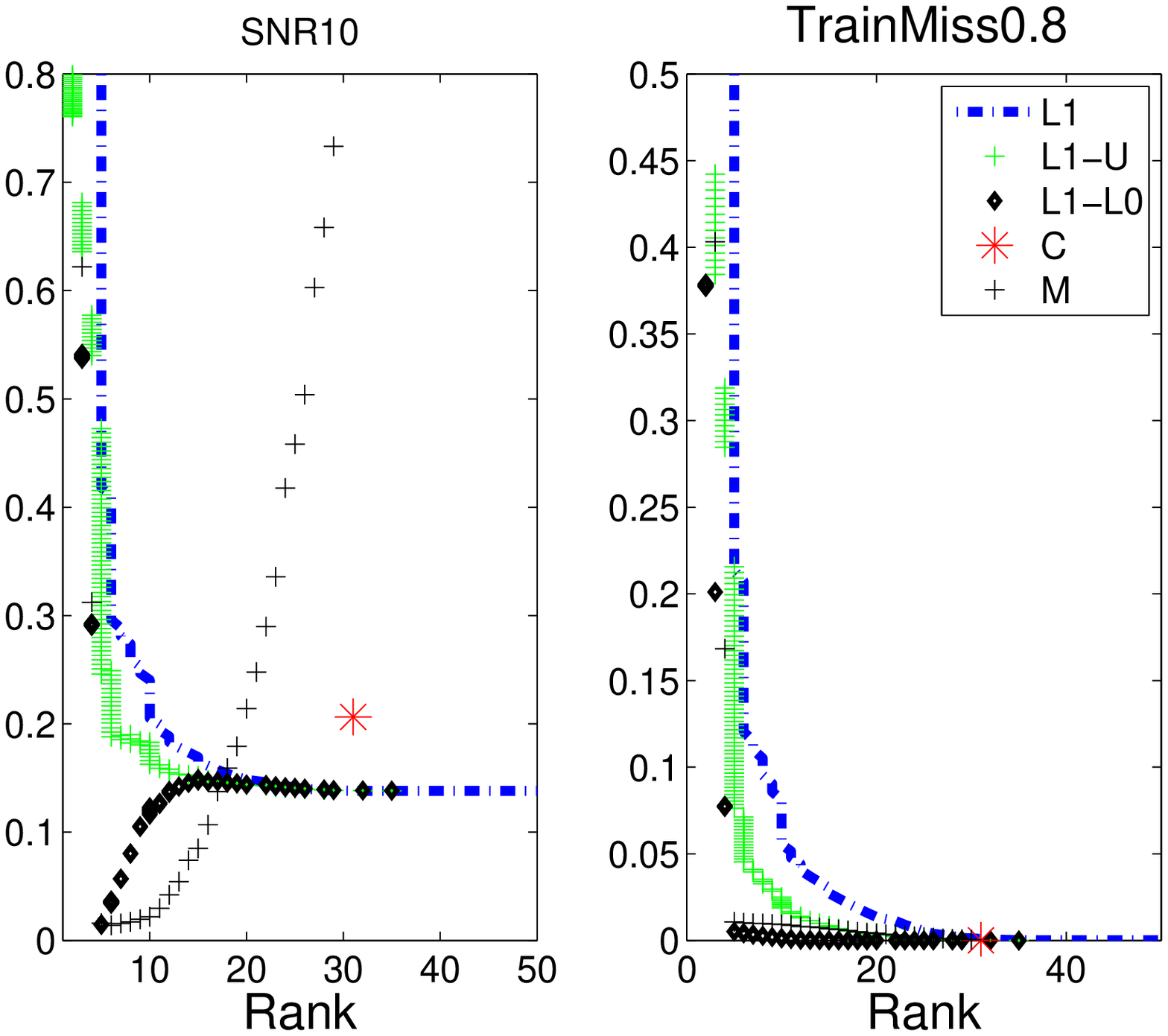} 
\end{psfrags}
	\caption{L1: solution for \textbf{Soft-Impute}; L1-U: Post
          processing after \textbf{Soft-Impute}; L1-L0
          \textbf{Hard-Impute} applied to L1-U; C : \textbf{SVT}
          algorithm; M: \textbf{Recon} algorithm. \textbf{Soft-Impute}
        performs well in the presence of noise (top and middle
        panel). When the noise is low, \textbf{Hard-Impute} can
        improve its performance.The post-processed version
tends to get the correct rank in many situations as in Types b,c.
In Type b, the post-processed version does better than the rest in prediction error.
In all the situations \textbf{SVT} algorithm does very poorly in
prediction error, confirming our claim that (\ref{crit:candes}) causes
overfitting.  \textbf{Recon} predicts poorly as well apart from Type-c, where it gets
better error than \textbf{Soft-Impute}. However \textbf{Hard-Impute} and
\textbf{Recon} have the same performance there.  
} 
\label{fig:eg-2}
\end{figure}

We believe that (\ref{crit:candes}) will almost always be too rigid,
as it will force the procedure to overfit. 
If minimization of prediction error is our main goal, then the
solution $Z^*$ will typically  lie somewhere in the interior of the path 
(Figure~\ref{fig:eg-2}), indexed by $\delta$.

In this paper we provide an algorithm for computing solutions of
(\ref{crit:relax}), on a grid of $\delta$ values, based on warm
restarts. The algorithm is inspired by Hastie et al.'s SVD- impute
\cite{svd-imp,olga01:_missin_dna} and is very different the proximal forward-backward
splitting method of \cite{cai-2008,pfbs1,fpc}, which requires the choice
of a step size. In \cite{fpc}, the SVD step becomes prohibitive,
so some randomized algorithms are used for the computation. Our
algorithm is very different, and by exploiting matrix structure
can solve problems much larger than those in \cite{fpc}.

Our algorithm requires the computation of a low-rank SVD of a matrix (which
is not sparse) at every iteration.  Here we crucially exploit the
problem matrix structure:
\begin{eqnarray}
Y= Y_{SP} \;\;(\mbox{Sparse})\quad +\quad Y_{LR}\;\;(\mbox{Low Rank})
\label{decomp}
\end{eqnarray}
In (\ref{decomp}) $Y_{SP}$ has the same sparsity structure as the
observed $X$, and $Y_{LR}$ has the rank $r \ll m,n$ of the estimated
$Z$. For large scale problems, we use iterative methods based on
Lanczos bidiagonalization with partial re-orthogonalization (as in the
PROPACK algorithm \cite{larsen98:_lancz}), for computing the first few singular
vectors/values of $Y.$ Due to the specific structure of
(\ref{decomp}), multiplication by $Y$ and $Y'$ can both be done in a
cost-efficient way..

\section{Algorithm and Convergence analysis}
\label{sec:nuc-norm-reg}
\subsection{Notation} \label{sec:notation}
We adopt the notation of \cite{cai-2008}. Define a matrix $P_{\Omega}(Y)$ (with dimension $n\times m$)
\begin{eqnarray} \label{notn:proj}
 P_{\Omega}(Y)\; (i,j) = \left\{\begin{array}{ll}
          Y_{i,j} & \mbox{if $(i,j) \in \Omega$}\\
          0 & \mbox{if $(i,j)  \notin \Omega$},\end{array} \right.
\end{eqnarray} 
which is a projection of the matrix $Y_{m\times n}$ onto the observed entries.
In the same spirit, define the complementary projection
$P^{\perp}_{\Omega}(Y)$ via 
$P^{\perp}_{\Omega}(Y)+P_{\Omega}(Y)=Y.$
 Using (\ref{notn:proj}) we can rewrite
$\sum_{(i,j) \in \Omega} (Z_{ij}-X_{ij})^2$ as $\|P_{\Omega}(Z)-P_{\Omega}(X)\|_F^2$.
\subsection{Nuclear norm regularization}
We present the following lemma, given in \cite{cai-2008},  which forms a basic ingredient in our algorithm. 
\begin{lem}
Suppose the matrix $W_{m\times n}$ has rank $r$. The solution to the
convex optimization problem
\begin{eqnarray}
\mini_Z \quad \half \|Z-W\|_F^2 + \lambda \|Z\|_*
\label{nuc-norm-basic}
\end{eqnarray} 
is given by $\hat W = \s_\lambda(W)$ where
\begin{eqnarray}
\s_\lambda(W) \equiv U D_\lambda V'\quad \mbox{ with } \quad D_\lambda=\mathrm{diag}\left[(d_1-\lambda)_+,\ldots,(d_r-\lambda)_+\right],
\label{svt}
\end{eqnarray}
where $X=UDV'$ is the SVD of $W$, 
$D=\mathrm{diag}\left[d_1,\ldots,d_r \right]$, and  $t_+=\max(t,0).$
\end{lem}
The notation $\s_\lambda(W)$ refers to {\em soft-thresholding} \cite{DJ95}.
The proof follows by looking at the sub-gradient of the function to be
minimized, and is given in \cite{cai-2008}. 

\subsection{Algorithm} \label{sec:algo}
Problem (\ref{crit:relax}) can be written in its equivalent Lagrangian form
\begin{equation}
\mini_Z \quad \half \|P_{\Omega}(Z)-P_{\Omega}(X)\|_F^2 + \lambda \|Z\|_* 
\label{crit:dual}
\end{equation}

Here $\lambda\geq 0$ is a regularization parameter controlling the
nuclear norm of the minimizer $ \hat{Z}_\lambda$ of
(\ref{crit:dual}) (with a 1-1 mapping to $\delta>0$ in (\ref{crit:relax})). 
We now present an algorithm for computing a series
of solutions to (\ref{crit:dual}) using warm starts.
Define
$f_\lambda(Z)= \half \|P_{\Omega}(Z)-P_{\Omega}(X)\|_F^2 + \lambda \|Z\|_*$.

\begin{algorithm}
\caption{\textbf{Soft-Impute}} \label{algo1}
\begin{enumerate}
\item Initialize $Z^{\mathrm{old}}=0$ and create a decreasing grid
  $\Lambda$ of values $\lambda_1>\ldots>\lambda_K$.
\item For every fixed $\lambda=\lambda_1,\;\lambda_2,\ldots\in\Lambda$
  iterate till convergence:
\begin{enumerate}
\item\label{item:1} Compute
$    Z^{\mathrm{new}}\leftarrow\s_\lambda(P_{\Omega}(X)+P_{\Omega}^{\perp}(Z^{\mathrm{old}}))$
 \item 
If $\quad\frac{\|f_\lambda(Z^{\mathrm{new}})-f_\lambda(Z^{\mathrm{old}})\|_F^2}{\|f_\lambda(Z^{\mathrm{old}})\|_F^2}< \epsilon,\quad$ go to step~\ref{item:2}.
\item Assign $Z^{\mathrm{old}}\leftarrow Z^{\mathrm{new}}$ and go to step~\ref{item:1}.
\item\label{item:2} Assign $\hat Z_\lambda\leftarrow Z^{\mathrm{new}}$ and  $Z^{\mathrm{old}}\leftarrow Z^{\mathrm{new}}$
\end{enumerate}
\item Output the sequence of solutions $\hat Z_{\lambda_1},\ldots,\hat Z_{\lambda_K}.$
\end{enumerate}
\end{algorithm}
The algorithm repeatedly replaces the missing entries with the current
guess, and then updates the guess by solving~(\ref{crit:dual}).
Figure~\ref{fig:eg-2} shows some examples of solutions using
Algorithm~\ref{algo1} (blue curves). We see test and training error in
the left two columns as a function of the nuclear norm, obtained from
a grid of values $\Lambda$. These error curves show a smooth and very
competitive performance.

\subsection{Convergence analysis} \label{conv-ana} 
In this section we prove that Algorithm~~\ref{algo1} converges to the
solution to~(\ref{crit:relax}).

For an arbitrary matrix $\tilde Z,$ define
\begin{equation}
Q_\lambda(Z|\tilde Z)=\half \|P_{\Omega}(X)+P_{\Omega}^{\perp}(\tilde Z)-Z\|_F^2 + \lambda \|Z\|_*,
\label{surr:defn}
\end{equation}
a surrogate of the objective function $f_\lambda(z)$. Note that 
$f_\lambda(\tilde Z)=Q_\lambda(\tilde Z|\tilde Z)$ for any $\tilde Z$.
\begin{lem}\label{sec:convergence-analysis}
For every fixed $\lambda\geq 0,$ define a sequence $Z_\lambda^k$ by
\begin{eqnarray} 
Z_\lambda^{k+1}&=&\arg\min_{Z} Q_\lambda(Z|Z_\lambda^k),
\label{surr}
\end{eqnarray}
with $Z_\lambda^0=0$.
The sequence $Z_\lambda^k$ satisfies 
\begin{eqnarray}
f_\lambda(Z_\lambda^{k+1})\leq Q_\lambda(Z_\lambda^{k+1}|Z_\lambda^k)\leq 
f_\lambda(Z_\lambda^k)
\label{wedge}
\end{eqnarray}
\end{lem}
\begin{proof}
\begin{eqnarray}
f_\lambda(Z_\lambda^k)&=& \half \|P_{\Omega}(X)+P_{\Omega}^{\perp}(Z_\lambda^k)-Z_\lambda^k\|_F^2 + \lambda \|Z_\lambda^k\|_* \nonumber\\
&\geq& \min_Z \{\|P_{\Omega}(X)+P_{\Omega}^{\perp}(Z_\lambda^k)-Z\|_F^2 + \lambda \|Z\|_* \}\nonumber\\
&=& Q_\lambda(Z_\lambda^{k+1}|Z_\lambda^k)  \nonumber\\
&=& \half \|\{P_{\Omega}(X)-P_{\Omega}(Z_\lambda^{k+1})\}\;+\{P_{\Omega}^{\perp}(Z_\lambda^k)-P_{\Omega}^{\perp}(Z_\lambda^{k+1})\}\|_F^2 + \lambda \|Z_\lambda^{k+1}\|_* \nonumber\\
&=& \half \; \{ \|P_{\Omega}(X)-P_{\Omega}(Z_\lambda^{k+1})\|_F^2 + 
\| P_{\Omega}^{\perp}(Z_\lambda^k)-P_{\Omega}^{\perp}(Z_\lambda^{k+1})\}\|_F^2 \} + \lambda \|Z_\lambda^{k+1}\|_* \nonumber\\
&\geq& \half \; \|P_{\Omega}(X)-P_{\Omega}(Z_\lambda^{k+1})\|_F^2 + \lambda \|Z_\lambda^{k+1}\|_* \nonumber\\
&=&Q_\lambda(Z_\lambda^{k+1}|Z_\lambda^{k+1}) \nonumber
\label{proof-monotone}
\end{eqnarray}
\end{proof}

\begin{lem}\label{lem:nonexpansive}
The nuclear norm shrinkage operator $\s_\lambda(\cdot)$ satisfies the following
for any $W_1,\;W_2$ (with matching dimensions)
\begin{eqnarray}
\|\s_\lambda(W_1)-\s_\lambda(W_2)\|_F^2 \leq \|W_1-W_2\|_F^2
\label{nonexpansive}
\end{eqnarray}
\end{lem}
\begin{proof}
  We omit the proof here for the sake of brevity. The details work out
  by expanding the operator $\s_\lambda(\cdot)$ in terms of the
  singular value decomposition of $W_1$ and $W_2.$ Then we use trace
  inequalities for the product of two matrices \cite{trace-ineq1}
  where one is real symmetric, the other arbitrary.  A proof of this
  Lemma also appears in \cite{fpc}, though the method is different
  from ours.
\end{proof}

\begin{lem}\label{lem:stationary}
Suppose the sequence $Z_\lambda^k$ obtained from (\ref{surr})
converges to $Z_\lambda^{\infty}.$ Then  $Z_\lambda^{\infty}$ is a stationary point of $f_\lambda(Z)$.
\end{lem}
\begin{proof}
The sub-gradients of the nuclear norm $\|Z\|_*$ are given by ~\cite{cai-2008}
\begin{eqnarray}
\partial \|Z\|_* =\{ UV' + W : W_{m\times n},\;U'W=0,\;WV=0,\; \|W\|_2 \leq 1 \}
\label{sub:nuc-norm}
\end{eqnarray}
where $Z=UDV'$ is the SVD of $Z$.
Since $Z_\lambda^k$ minimizes $Q_\lambda(Z|Z_\lambda^{k-1})$,
it satisfies:
\begin{eqnarray}
0 \in -(P_{\Omega}(X)+P_{\Omega}^{\perp}(Z_\lambda^{k-1})-Z_\lambda^k) + \partial \|Z_\lambda^k\|_*\;\;\forall k
\label{surr:stationary}
\end{eqnarray}
Since $Z_\lambda^k \rightarrow Z_\lambda^{\infty},$ 
\begin{eqnarray}
(P_{\Omega}(X)+P_{\Omega}^{\perp}(Z_\lambda^{k-1})-Z_\lambda^k) \longrightarrow 
(P_{\Omega}(X)-P_{\Omega}(Z_\lambda^{\infty})).
\label{lim:one}
\end{eqnarray}
For every $k,$ a sub-gradient $p(Z_\lambda^k) \in \partial \|Z_\lambda^k\|_*$ corresponds to a tuple $(u_{k},v_{k},w_{k}).$ 
 Then (passing on to a subsequence if necessary),
$(u_{k},v_{k},w_{k})\rightarrow (u_{\infty},v_{\infty},w_{\infty})$ and this limit corresponds to 
$p(Z_\lambda^{\infty})\in \partial \|Z_\lambda^{\infty}\|_*$.

Hence, from (\ref{surr:stationary}, \ref{lim:one}), passing on to the limits
\begin{eqnarray}
\mathbf{0} \in (P_{\Omega}(X) - P_{\Omega}(Z_\lambda^{\infty})) + \partial \|Z_\lambda^{\infty}\|_*
\label{lim:infty}
\end{eqnarray}
This proves the stationarity of the limit $Z_\lambda^{\infty}$.
\end{proof}

\begin{thm}
The sequence $Z_\lambda^k$ defined in Lemma~\ref{sec:convergence-analysis}
converges to $Z_\lambda^{\infty}$ which solves 
\begin{eqnarray}
\min_Z \half \|P_{\Omega}(Z)-P_{\Omega}(X)\|_F^2 + \lambda \|Z\|_*  
\label{fixed:pt}
\end{eqnarray}
\end{thm}
\begin{proof}
Firstly observe that the sequence  $Z_\lambda^k$ is bounded; for it to converge it must have a unique accumulation point. 

Observe that 
\begin{eqnarray}
\|Z_\lambda^{k+1}-Z_\lambda^k\|_F^2&=&
 \|\s_\lambda(P_{\Omega}(X)+P_{\Omega}^{\perp}(Z_\lambda^k))-\s_\lambda(P_{\Omega}(X)+P_{\Omega}^{\perp}(Z_\lambda^{k-1}))\|_F^2  \nonumber\\
(\mbox{by Lemma~\ref{lem:nonexpansive}})&\leq&\|\left(P_{\Omega}(X)+P_{\Omega}^{\perp}(Z_\lambda^k)\right)-\left(P_{\Omega}(X)+P_{\Omega}^{\perp}(Z_\lambda^{k-1})\right)\|_F^2 \nonumber \\
&=&\|P_{\Omega}^{\perp}(Z_\lambda^k-Z_\lambda^{k-1})\|_F^2 \nonumber\\
&\leq&\|Z_\lambda^k-Z_\lambda^{k-1}\|_F^2  \label{contrac}
\end{eqnarray}

Due to boundedness, every infinite subsequence of $Z_\lambda^k$ has a
further subsequence that converges.  If the sequence $Z_\lambda^k$ has
two distinct limit points then for infinitely many $k'\geq 0,$
$\|Z_\lambda^{k'}-Z_\lambda^{k'-1}\|_F\geq \epsilon,$ for some
$\epsilon>0$.  Using (\ref{contrac}) this contradicts the convergence
of any subsequence of $Z_\lambda^k.$ Hence the sequence $Z_\lambda^k$
converges.  Using Lemma~\ref{lem:stationary}, the limit
$Z_\lambda^{\infty}$ is a stationary point of $f_\lambda(Z)$ and hence
its minimizer.
\end{proof}

\section{From soft to hard-thresholding} \label{sec:soft-hard} The
nuclear norm behaves like a $\ell_1$ norm, and can be viewed as a soft
approximation of the $\ell_0$ norm or rank of a matrix.  In penalized
linear regression for example, the $\ell_1$ norm or LASSO \cite{Ti96}
is widely used as a convex surrogate for the $\ell_0$ penalty or
best-subset selection. The LASSO performs very well on a wide variety
of situations in producing a parsimonious model with good prediction
error. However, if the underlying model is very sparse, then the LASSO
with its uniform shrinkage can overestimate the number of non-zero
coefficients.  In such situations concave penalized regressions are
gaining popularity as a surrogate to $\ell_0$. By analogy for matrices, it makes
sense to go beyond the nuclear norm minimization problem to more
aggressive penalties bridging the gap between $\ell_1$ and $\ell_0$.
We propose minimizing
\begin{eqnarray}
f_{p,\lambda}(Z)&=& \half \|P_{\Omega}(Z)-P_{\Omega}(X)\|_F^2 + \lambda \sum_j p(\lambda_j(Z);\gamma)
\label{crit:relax-concave}
\end{eqnarray}
where $p(|t|;\gamma)$ is concave in $|t|.$ The parameter $\gamma \in
[\gamma_{\inf},\gamma_{\sup}]$ controls the degree of concavity, with
$p(|t|;\gamma_{\inf})=|t|$ ($\ell_1$ penalty), on one end and
$p(|t|;\gamma_{\sup})=|t|^0$ ($\ell_0$ penalty) on the other. In
particular for the $\ell_0$ penalty denote $f_{p,\lambda}(Z)$ by
$f_{H,\lambda}(Z)$ for ``hard'' thresholding. See
\cite{FJ08,Fan01,Zhang07} for examples of such penalties.

Criterion (\ref{crit:relax-concave}) is no longer convex and hence
becomes more difficult.  It can be shown that Algorithm~\ref{algo1}
can be modified in a suitable fashion for the penalty
$p(\cdot;\gamma).$ This algorithm also has guaranteed convergence
properties.
The details of these arguments and statistical properties will be
studied in a longer version of this paper.  As a concrete example, we
present here some features of the $\ell_0$ norm regularization on
singular values.

The version of (\ref{nuc-norm-basic})  for the $\ell_0$ norm is
\begin{eqnarray}
\min_Z \half \|Z-W\|_F^2 + \lambda \|Z\|_0.
\label{l0-norm-basic}
\end{eqnarray} 
The solution is given by a reduced-rank SVD of $W$; for every
$\lambda$ there is a corresponding $q=q(\lambda)$ number of singular-values to be
retained in the SVD decomposition.  As in (\ref{svt}), the
thresholding operator resulting from (\ref{l0-norm-basic}) is
\begin{eqnarray}
\s^H_\lambda(W) = U D_q V'\quad \mathrm{where} \quad D_{q}=\mathrm{diag}\left(d_1,\ldots,d_q,0,\ldots,0\right)
\label{svt-hard}
\end{eqnarray}
 
Similar to \textbf{Soft-Impute} (Algorithm \ref{algo1}), 
the algorithm \textbf{Hard-Impute} for the $\ell_0$ penalty is given by Algorithm \ref{algo2}.
\begin{algorithm}
\caption{\textbf{Hard-Impute}}\label{algo2}

\begin{enumerate}
\item Create a decreasing grid
  $\Lambda$ of values $\lambda_1>\ldots>\lambda_K$. Initialize
  $\tilde Z_{\lambda_k}\;k=1,\ldots,K$ (see Section~\ref{pp-and-init}).
\item For every fixed $\lambda=\lambda_1,\;\lambda_2,\ldots\in\Lambda$
  iterate till convergence:
\begin{enumerate}
\item Initialize $Z^{\mathrm{old}}\leftarrow \tilde Z_\lambda$.
\item\label{item:1} Compute
$    Z^{\mathrm{new}}\leftarrow\s_\lambda^H(P_{\Omega}(X)+P_{\Omega}^{\perp}(Z^{\mathrm{old}}))$
 \item If
$\quad\frac{\|f_\lambda(Z^{\mathrm{new}})-f_\lambda(Z^{\mathrm{old}})\|_F^2}{\|f_\lambda(Z^{\mathrm{old}})\|_F^2}< \epsilon,\quad$ 
go to step~\ref{item:2}.
\item Assign $Z^{\mathrm{old}}\leftarrow Z^{\mathrm{new}}$ and go to step~\ref{item:1}.
\item\label{item:2} Assign $\hat Z_{H,\lambda}\leftarrow Z^{\mathrm{new}}$.
\end{enumerate}
\item Output the sequence of solutions $\hat Z_{H,\lambda_1},\ldots,\hat Z_{\lambda_K}.$
\end{enumerate}
\end{algorithm}

\subsection{Post-processing and Initialization} \label{pp-and-init}
Because the $\ell_1$ norm regularizes by shrinking the singular
values, the number of singular values retained (through
cross-validation, say) may exceed the actual rank of the matrix.  In
such cases it is reasonable to {\em undo} the shrinkage of the chosen
models, which might permit a lower-rank solution.

If $Z_\lambda$ is the solution to~(\ref{crit:dual}), then its
\emph{post-processed} version $Z^u_\lambda$ 
obtained by ``unshrinking'' the eigen-values of the matrix $Z_\lambda$
is obtained by
\begin{eqnarray}
  \alpha&=& \argmin_{\alpha_i\geq 0,\;i=1,\ldots,r_\lambda} \quad \|P_{\Omega}(X)-\sum_{i=1}^{r_\lambda} \alpha_i P_{\Omega}(u_iv_i')\|^2 \label{post-process}\\
  Z^u_\lambda&=& UD_\alpha V',\nonumber
\end{eqnarray}   
where $D_\alpha=\mbox{diag}(\alpha_1,\ldots,\alpha_{r_\lambda})$.
Here $r_\lambda$ is the rank of $Z_\lambda$ and $Z_\lambda=UD_\lambda
V'$ is its SVD.  The estimation in (\ref{post-process}) can be done via ordinary
least squares, which is feasible because of the sparsity of
$P_{\Omega}(u_iv_i')$ and that  $r_\lambda$ is small.\footnote{Observe that
the   $P_{\Omega}(u_iv_i'),\; i=1,\ldots,r_\lambda$  are not orthogonal, though
the   $u_iv_i'$ are.} 
If the least squares solutions
$\boldsymbol\alpha$ do not meet the positivity constraints, then the
negative sign can be absorbed into the corresponding singular vector.

In many simulated examples we have observed that this post-processing
step gives a good estimate of the underlying true rank of the matrix
(based on prediction error).  Since fixed points of
Algorithm~\ref{algo2} correspond to local minima of the
function~(\ref{crit:relax-concave}), well-chosen warm starts $\tilde
Z_\lambda$ are helpful.  A reasonable prescription for warms-starts is
the nuclear norm solution via (\textbf{Soft-Impute}), or the post
processed version~(\ref{post-process}). The latter appears to 
significantly speed up convergence for \textbf{Hard-Impute}.

\subsection{Computation} \label{comp} The computationally demanding
part of Algorithms~\ref{algo1} and \ref{algo2} is in
$\s_\lambda(P_{\Omega}(X)+P_{\Omega}^{\perp}(Z_\lambda^k))$ or
$\s^{H}_\lambda(P_{\Omega}(X)+P_{\Omega}^{\perp}(Z_{H,\lambda}^k))$.
These require calculating a low- rank SVD of the matrices of
interest, since the underlying model assumption is that $\rnk(Z) \ll
\min\{m,n\}$.  In Algorithm~\ref{algo1}, for fixed $\lambda,$ the entire sequence of matrices
$Z_\lambda^k$ have explicit low-rank representations of the form $U_k
D_k V'_k$ corresponding to
$\s_\lambda(P_{\Omega}(X)+P_{\Omega}^{\perp}(Z_\lambda^{k-1}))$

In addition, observe that $P_{\Omega}(X)+P_{\Omega}^{\perp}(Z_\lambda^k)$ can be rewritten as  
\begin{eqnarray}
P_{\Omega}(X)+P_{\Omega}^{\perp}(Z_\lambda^k) = \left\{P_{\Omega}(X)-P_{\Omega}(Z_\lambda^k)\right\}\; (\mathrm{Sparse}) \quad +
\quad Z_\lambda^k \; (\mathrm{Low Rank})
\label{sparse-low-rank-upd}
\end{eqnarray}

In the numerical linear algebra literature, there are very efficient
direct matrix factorization methods for calculating the SVD of
matrices of moderate size (at most a few thousand). When the matrix is
sparse, larger problems can be solved but the computational cost
depends heavily upon the sparsity structure of the matrix.  In general
however, for large matrices one has to resort to indirect iterative
methods for calculating the leading singular vectors/values of a
matrix. There is a lot research in the numerical linear algebra for
developing sophisticated algorithms for this purpose.  In this paper
we will use the PROPACK algorithm \cite{propack,larsen98:_lancz}
because of its low storage requirements, effective flop count and its
well documented MATLAB version.  The algorithm for calculating the
truncated SVD for a matrix $W$ (say), becomes
efficient if multiplication operations $Wb_1$ and $W'b_2$
(with $b_1 \in \Re^n,\;b_2 \in \Re^m$) can be done with minimal cost.

Our algorithms \textbf{Soft-Impute} and \textbf{Hard-Impute} both
require repeated computation of a truncated SVD for a matrix 
$W$ with structure as in (\ref{sparse-low-rank-upd}). 
Note that in (\ref{sparse-low-rank-upd}) the term
$P_{\Omega}(Z_\lambda^k)$ can be computed in  
 $O(|\Omega|r)$ flops using only the required  outer products.

 The cost of computing the truncated SVD will depend upon the cost in
 the operations $Wb_1$ and $W'b_2$ (which are equal). For the sparse
 part these multiplications cost $O(|\Omega|)$. Although it costs 
 $O(|\Omega|r)$ to create the  matrix $P_{\Omega}(Z_\lambda^k)$), this
 is used for each of the $r$ such multiplications (which also cost
 $O(|\Omega|r)$), so we need not include that cost here.
 The $\mathrm{Low Rank}$ part costs $O((m+n)r)$ for the multiplication
 by $b_1$.  Hence the cost is $O(|\Omega|)+O((m+n)r)$ per multiplication. 
 cost.

 For the reconstruction problem to be theoretically meaningful in the
 sense of ~\cite{candes-2009}, we require that $|\Omega|\approx
 nr\mathrm{poly}(\log n).$ Hence introducing the $\mathrm{Low Rank}$
 part does not add any further complexity in the multiplication by $W$
 and $W'$. So the dominant cost in calculating the truncated SVD in
 our algorithm is $O(|\Omega|)$.  The \textbf{SVT} algorithm
 \cite{cai-2008} for exact matrix completion (\ref{crit:candes})
 involves calculating the SVD of a sparse matrix with cost
 $O(|\Omega|).$ This implies that the computational cost
 of our algorithm and that of
 ~\cite{cai-2008} is the same.  Since the true rank of the matrix
 $r\ll \min\{m,n\},$ the computational cost of evaluating the
 truncated SVD (with rank $\approx r$) is linear in matrix dimensions.
 This justifies the large-scale computational feasibility of our
 algorithm.
 
 The PROPACK package does not allow one to request (and hence compute)
 only the singular values larger than a threshold $\lambda$ --- one
 has to specify the number in advance. So once all the computed
 singular values fall above the current threshold $\lambda$, our
 algorithm increases the number to be computed until the smallest is
 smaller than $\lambda$. In large scale problems, we put an absolute
 limit on the maximum number.

 \section{Simulation Studies} \label{sec:simu} In this section we
 study the training and test errors achieved by the estimated matrix
 by our proposed algorithms and those by \cite{cai-2008,monti-09}. The
 Reconstruction algorithm (\textbf{Rcon}) described in \cite{monti-09}
 considers criterion~(\ref{crit:one}) (in presence of noise).
For every fixed rank $r$ it uses a bi-convex algorithm on a
 Grassmanian Manifold for computing a rank-$r$ approximation $USV'$
 (not the SVD). It uses a suitable starting point obtained
 by performing a sparse SVD on a \emph{clean} version of the observed
 matrix $P_{\Omega}(X).$ To summarize, we look at the performance
 of the following methods:
\begin{itemize}
\item (a) \textbf{Soft-Impute} (algorithm \ref{algo1});  (b) Post-processing on the output of Algorithm \ref{algo1},
(c) \textbf{Hard-Impute} (Algorithm \ref{algo2}) starting with the output of (b).
\item \textbf{SVT} algorithm by \cite{cai-2008} 
\item \textbf{Rcon} reconstruction algorithm by \cite{monti-09}
\end{itemize}
In all our simulation studies we took the underlying model as
$Z_{m\times n}=U_{m \times r} V'_{r \times n} + \mathrm{noise};$ where
$U$ and $V$ are random matrices with standard normal Gaussian entries,
and $\mathrm{noise}$ is iid Gaussian. $\Omega$ is uniformly random
over the indices of the matrix with $p\%$ percent of missing entries.
These are the models under which the coherence conditions hold true
for the matrix completion problem to be meaningful as pointed out in
~\cite{candes-2009,monti-09}. The signal to noise ratio for the model
and the test-error (standardized) are defined as
\begin{eqnarray}
\mathrm{SNR}=\sqrt{\frac{\mbox{var}(UV')}{\mbox{var}(noise)}};\quad 
\mathrm{testerror}=\frac{\|P_{\Omega}^{\perp}(UV'-\hat Z)\|_F^2}{\|P_{\Omega}^{\perp}(UV')\|_F^2}
\end{eqnarray}
In Figure~\ref{fig:eg-2}, results corresponding to the training and
test errors are shown for all algorithms mentioned above --- 
nuclear norm (left two panels) and rank (right two panels)--- in three
problem instances. Since  \textbf{Rcon} only uses rank, it is excluded
from the left panels.
In all examples
$(m,n)=(100,100).$ SNR, true rank and percentage of missing
entries are indicated in the figures. There is a unique correspondence
between $\lambda$ and nuclear norm. The plots vs the rank indicate
how effective the nuclear norm is as a rank approximation --- that is
whether it recovers the true rank while minimizing prediction
error. We summarize our findings in the caption of the figure.

In addition we performed some large scale simulations in
Table~\ref{tab:one} for our algorithm in different problem sizes. The
problem dimensions, SNR, number of iterations till convergence and
time in seconds are reported. All computations are done in MATLAB and
the MATLAB version of PROPACK is used.

\subsubsection*{Acknowledgements}
\label{sec:acknowledgements}
We thank Emmanuel Candes, Andrea Montanari and Steven Boyd for helpful discussions. Trevor Hastie was partially supported by grant DMS-0505676 from the National
Science Foundation, and grant 2R01 CA 72028-07 from the National Institutes of
Health.



\begin{table}
 \begin{tabular}[ht!!]{c|c|c|c|c|c|c}  \\
$(m,n)$  &         $|\Omega|$ & true rank ($r$) & SNR& effective rank ($\hat r$) & \# Iters & time(s)\\\hline
$(3 \times 10^4,10^4)$&     $10^4$    &   $15$  &  $1$ &  $(13,47,80)$ & $(3,3,3)$ & $(41.9,124.7,305.8)$\\
$(5\times 10^4,5 \times 10^4)$&  $10^4$ &     $15$ &  $1$  &  $8$ &   $80$ &   $237$\\
$(10^5,10^5)$   &  $10^4$   &      $15$ &    $10$ &   $(5,14,32,62)$ & $(3,3,3,3)$ & $(37,74.5,199.8,653)$\\
$(10^5,10^5)$   & $10^5$    &     $15$ &    $10$ & $(18,80)$ & $(3,3)$ & $(202, 1840)$\\
 $(5\times10^5,5\times10^5)$    & $10^4$   &     $15$ &    $10$ &  $11$ &  $ 3$  & $628.14$ \\
 $(5\times10^5,5\times10^5)$    & $10^5$   &     $15$ &    $1$ &  $(3,11,52)$ &  $(3,3,3)$  & $(341.9,823.4,4810.75)$ \\\hline
\end{tabular}
\caption{Performance of the \textbf{Soft-Impute} on different problem instances.} \label{tab:one}
\end{table}

\newpage
\bibliographystyle{alpha}
\bibliography{/home/rahulm/rahul/Trevor/Mat_completion/mrc.bib,/home/hastie/bibtex/tibs.bib,/home/hastie/docs/resume/trevor.bib}

\end{document}